# A Multi-threshold Segmentation Approach Based on Artificial Bee Colony Optimization

**Erik Cuevas, Felipe Sención-Echauri, Daniel Zaldivar[1] and Marco Pérez-Cisneros**

Departamento de Ciencias Computacionales
Universidad de Guadalajara, CUCEI
Av. Revolución 1500, Guadalajara, Jal, México
{erik.cuevas, felipe.sencion, [1]daniel.zaldivar, marco.perez}@cucei.udg.mx

**Abstract**

This paper explores the use of the Artificial Bee Colony (ABC) algorithm to compute threshold selection for image segmentation. ABC is a heuristic algorithm motivated by the intelligent behavior of honey-bees which has been successfully employed to solve complex optimization problems. In this approach, an image 1-D histogram is approximated through a Gaussian mixture model whose parameters are calculated by the ABC algorithm. For the approximation scheme, each Gaussian function represents a pixel class and therefore a threshold. Unlike the Expectation-Maximization (EM) algorithm, the ABC-based method shows fast convergence and low sensitivity to initial conditions. Remarkably, it also improves complex time-consuming computations commonly required by gradient-based methods. Experimental results demonstrate the algorithm's ability to perform automatic multi-threshold selection yet showing interesting advantages by comparison to other well-known algorithms.

## 1. Introduction

Several image processing applications aim to detect and classify relevant features which may be later analyzed to perform several high-level tasks. In particular, image segmentation seeks to group pixels within meaningful regions. Commonly, gray levels belonging to the object, are substantially different from those featuring the background. Thresholding is thus a simple but effective tool to isolate objects of interest; its applications include several classics such as document image analysis, whose goal is to extract printed characters [1,2], logos, graphical content, or musical scores; also it is used for map processing which aims to locate lines, legends, and characters [3]. Moreover, it is employed for scene processing, seeking for object detection, marking [4] and for quality inspection of materials [5,6].

Thresholding selection techniques can be classified into two categories: bi-level and multi-level. In the former, one limit value is chosen to segment an image into two classes: one representing the object and the other one segmenting the background. When distinct objects are depicted within a given scene, multiple threshold values have to be selected for proper segmentation, which is commonly called multilevel thresholding.

A variety of thresholding approaches have been proposed for image segmentation, including conventional methods [7, 8, 9, 10] and intelligent techniques [11, 12]. Extending the segmentation algorithms to a multilevel approach may cause some inconveniences: (i) they may have no systematic or analytic solution when the number of classes to be detected increases and (ii) they may also show a slow convergence and/or high computational cost [11].

In this work, the segmentation algorithm is based on a parametric model holding a probability density function of gray levels which groups a mixture of several Gaussian density functions (Gaussian mixture). Mixtures represent a flexible method of statistical modelling as they are employed in a wide variety of contexts [13]. Gaussian mixture has received considerable attention in the development of segmentation algorithms despite its performance is influenced by the shape of the image histogram and the accuracy of the estimated model parameters [14]. The associated parameters can be calculated considering the Expectation Maximization (EM) algorithm [15,16] or Gradient-based methods such as Levenberg-Marquardt, LM [17]. However, EM algorithms are very sensitive to the choice of the initial values [18], meanwhile Gradient-based methods are computationally expensive and may easily get stuck within local minima [14]. Therefore, some researchers have attempted to develop methods based on modern global optimization algorithms such as the Learning Automata (LA) [19] and differential evolution algorithm [20]. In this paper, an alternative approach using an optimization algorithm for determining the parameters of a Gaussian mixture is presented.

[1] Corresponding author, Tel +52 33 1378 5900, ext. 7715, E-mail: daniel.zaldivar@cucei.udg.mx

On other hand, Karaboga [21] has presented a metaheuristic algorithm for solving numerical optimization problems known as the artificial bee colony (ABC) method. Inspired by the intelligent foraging behavior of a honeybee swarm, the ABC algorithm consists of three essential components: food source positions, nectar-amounts and several honey-bee classes. Each food source position represents a feasible solution for the problem under consideration. The nectar-amount for a food source represents the quality of such solution according to its fitness value. Each bee-class symbolizes one particular operation for generating new candidate food source positions (i.e. candidate solutions).

The ABC algorithm starts by producing a randomly distributed initial population (food source locations). After initialization, an objective function evaluates whether such candidates represent an acceptable solution (nectar-amount) or not. Guided by the values of such objective function, candidate solutions are evolved through different ABC operations (honey-bee types). When the fitness function (nectar-amount) cannot be further improved after a maximum number of cycles, its related food source is assumed to be abandoned and replaced by a new randomly chosen food source location.

The performance of ABC algorithm has been compared to other metaheuristic methods such as Genetic Algorithms (GA), Differential Evolution (DE) and Particle Swarm Optimization (PSO) [22,23]. The results have shown that ABC can produce optimal solutions yet more effectively than other methods for several optimization problems. Such characteristics have motivated the use of ABC to solve different sorts of engineering problems within different fields such as signal processing [24], flow shop scheduling [25], structural inverse analysis [26], clustering [27,28] and electromagnetism [29].

This paper presents the use of the Artificial Bee Colony (ABC) algorithm to compute threshold selection for image segmentation. In this approach, the segmentation process is considered as an optimization problem approximating the 1-D histogram of a given image by means of a Gaussian mixture model. The operation parameters are calculated through the ABC algorithm. Each Gaussian function approximating the histogram represents a pixel class and therefore a threshold point in the segmentation scheme. The experimental results, presented in this work, demonstrate that ABC exhibits fast convergence, relative low computational cost and no sensitivity to initial conditions by keeping an acceptable segmentation of the image, i.e. a better mixture approximation in comparison to the EM or gradient based algorithms.

The paper is organized as follows: Section 2 presents the Gaussian approximation of the histogram while Section 3 discusses on the ABC algorithm. Section 4 formulates the threshold determination with Section 5 presenting all experimental results after the proposed approach is implemented. Section 6 summarizes a full discussion on the algorithm performance.

## 2. Gaussian approximation

Let consider an image holding $L$ gray levels $[0,\ldots,L-1]$ whose distribution is displayed within a histogram $h(g)$. In order to simplify the description, the histogram is normalized just as a probability distribution function, yielding:

$$h(g)=\frac{n_g}{N},\quad h(g)>0, \tag{1}$$

$$N = \sum_{g=0}^{L-1} n_g, \text{ and } \sum_{g=0}^{L-1} h(g) = 1,$$

where $n_g$ denotes the number of pixels with gray level $g$ and $N$ being the total number of pixels in the image. The histogram function can thus be contained into a mix of Gaussian probability functions of the form:

$$p(x) = \sum_{i=1}^{K} P_i \cdot p_i(x) = \sum_{i=1}^{K} \frac{P_i}{\sqrt{2\pi\sigma_i}} \exp\left[\frac{-(x-\mu_i)^2}{2\sigma_i^2}\right] \tag{2}$$

with $P_i$ being the probability of class $i$, $p_i(x)$ being the probability distribution function of gray-level random variable $x$ in class $i$, with $\mu_i$ and $\sigma_i$ being the mean and standard deviation of the $i$-th probability distribution function and $K$ being the number of classes within the image. In addition, the constraint $\sum_{i=1}^{K} P_i = 1$ must be satisfied.

The mean square error is used to estimate the $3K$ parameters $P_i$, $\mu_i$ and $\sigma_i$, $i$ = 1, . . ,$K$. For instance, the mean square error between the Gaussian mixture $p(x_i)$ and the experimental histogram function $h(x_i)$ is defined as follows:

$$J = \frac{1}{n}\sum_{j=1}^{n}\left[p(x_j) - h(x_j)\right]^2 + \omega \cdot \left|\left(\sum_{i=1}^{K} P_i\right) - 1\right| \tag{3}$$

Assuming an $n$-point histogram as in [13] and $\omega$ being the penalty associated with the constrain $\sum_{i=1}^{K} P_i = 1$. In general, the parameter estimation that minimizes the square error produced by the Gaussian mixture is not a simple problem. A straightforward method is to consider the partial derivatives of the error function to zero by obtaining a set of simultaneous transcendental equations [13]. However, an analytical solution is not always available considering the non-linear nature of the equation which in turn yields the use of iterative approaches such as gradient-based or maximum likelihood estimation. Unfortunately, such methods may also get easily stuck within local minima or be time expensive.

In the case of other algorithms such as the EM algorithm and the gradient-based methods, the new parameter point lies within a neighbourhood distance of the previous parameter point. However, this is not the case for the ABC adaptation algorithm which is based on stochastic principles. New operating points are thus determined by a parameter probability function that may yield points lying far away from previous operating points, providing the algorithm with a higher ability to locate and pursue a global minimum.

## 3. Artificial Bee Colony (ABC) algorithm

The ABC algorithm assumes the existence of a set of operations that may resemble some features of the honey bee behavior. For instance, each solution within the search space includes a parameter set representing food source locations. The "fitness value" refers to the food source quality that is strongly linked to the food's location. The process mimics the bee's search for valuable food sources yielding an analogous process for finding the optimal solution.

### *3.1 Biological bee profile*

The minimal model for a honey bee colony consists of three classes: employed bees, onlooker bees and scout bees. The employed bees will be responsible for investigating the food sources and sharing the information with recruit onlooker bees. They, in turn, will make a decision on choosing food sources by considering such information. The food source having a higher quality will have a larger chance to be selected by onlooker bees than those showing a lower quality. An employed bee, whose food source is rejected as low quality by employed and onlooker bees, will change to a scout bee to randomly search for new food sources. Therefore, the exploitation is driven by employed and onlooker bees while the exploration is maintained by scout bees. The implementation details of such bee-like operations in the ABC algorithm are described in the next sub-section.

### *3.2 Description of the ABC algorithm*

Resembling other metaheuristic approaches, the ABC algorithm is an iterative process. It starts with a population of randomly generated solutions or food sources. The following three operations are applied until a termination criterion is met [23]:

1. Send the employed bees.
2. Select the food sources by the onlooker bees.
3. Determine the scout bees.

#### *3.2.1 Initializing the population*

The algorithm begins by initializing $N_p$ food sources. Each food source is a *D*-dimensional vector containing the parameter values to be optimized, which are randomly and uniformly distributed between the pre-specified lower initial parameter bound $x_j^{low}$ and the upper initial parameter bound $x_j^{high}$.

$$x_{j,i} = x_j^{low} + \text{rand}(0,1)\cdot(x_j^{high} - x_j^{low}); \quad j = 1,2,\ldots,D; \quad i = 1,2,\ldots,N_p. \tag{4}$$

with *j* and *i* being the parameter and individual indexes respectively. Hence, $x_{j,i}$ is the *j*th parameter of the *i*th individual.

#### *3.2.2 Send employed bees*

The number of employed bees is equal to the number of food sources. At this stage, each employed bee generates a new food source in the neighborhood of its present position as follows:

$$v_{j,i} = x_{j,i} + \phi_{j,i}(x_{j,i} - x_{j,k}); \quad k \in \{1,2,\ldots,N_p\}; j \in \{1,2,\ldots,D\} \tag{5}$$

$x_{j,i}$ is a randomly chosen *j* parameter of the *i*th individual and *k* is one of the $N_p$ food sources, satisfying the condition $i \neq k$. If a given parameter of the candidate solution $v_i$ exceeds its predetermined boundaries, that parameter should be adjusted in order to fit the appropriate range. The scale factor $\phi_{j,i}$ is a random number between $[-1,1]$. Once a new solution is generated, a fitness value representing the profitability associated with a particular solution is calculated. The fitness value for a minimization problem can be assigned to each solution $v_i$ by the following expression:

$$fit_i = \begin{cases} \dfrac{1}{1+J_i} & \text{if } J_i \geq 0 \\ 1 + abs(J_i) & \text{if } J_i < 0 \end{cases} \tag{6}$$

where $J_i$ is the objective function to be minimized. A greedy selection process is thus applied between $v_i$ and $x_i$. If the nectar- amount (fitness) of $v_i$ is better, then the solution $x_i$ is replaced by $v_i$; otherwise, $x_i$ remains.

*3.2.3 Select the food sources by the onlooker bees*

Each onlooker bee (the number of onlooker bees corresponds to the food source number) selects one of the proposed food sources, depending on their fitness value, which has been recently defined by the employed bees. The probability that a food source will be selected can be obtained from the following equation:

$$Prob_i = \frac{fit_i}{\sum_{i=1}^{N_p} fit_i} \tag{7}$$

where $fit_i$ is the fitness value of the food source *i*, which is related to the objective function value ($J_i$) corresponding to the food source *i*. The probability of a food source being selected by onlooker bees increases with an increase in the fitness value of the food source. After the food source is selected, onlooker bees will go to the selected food source and select a new candidate food source position inside the neighborhood of the selected food source. The new candidate food source can be expressed and calculated by Eq.(5). In case the nectar-amount, i.e., fitness of the new solution, is better than before, such position is held; otherwise, the last solution remains.

*3.2.4 Determine the scout bees*

If a food source *i* (candidate solution) cannot be further improved through a predetermined trial number known as "limit", the food source is assumed to be abandoned and the corresponding employed or onlooker bee becomes a scout. A scout bee explores the searching space with no previous information, i.e., the new solution is generated randomly as indicated by Eq.(4). In order to verify if a candidate solution has reached the predetermined "*limit*", a counter $A_i$ is assigned to each food source *i*. Such a counter is incremented consequent to a bee-operation failing to improve the food source's fitness.

## 4. Determination of Thresholding Values

In order to determine optimal threshold values, it is considered that the data classes are organized such that $\mu_1 < \mu_2 < \ldots < \mu_K$. Therefore, threshold values are obtained by computing the overall probability error of two adjacent Gaussian functions, yielding:

$$\begin{gathered} E(T_h) = P_{h+1} \cdot E_1(T_h) + P_i \cdot E_2(T_h), \\ h = 1, 2, \ldots, K-1 \end{gathered} \tag{8}$$

considering

$$E_1(T_h) = \int_{-\infty}^{T_h} p_{h+1}(x)dx, \tag{9}$$

and

$$E_2(T_h) = \int_{T_h}^{\infty} p_h(x)dx, \tag{10}$$

$E_1(T_h)$ is the probability of mistakenly classifying the pixels in the ($h$ + 1)-th class belonging to the $h$-th class, while $E_2(T_h)$ is the probability of erroneously classifying the pixels in the $h$-th class belonging to the ($h$ + 1)-th class. $P_j'$s are the a priori probabilities within the combined probability density function, and $T_h$ is the threshold value between the $h$-th and the ($h$ + 1)-th classes. One $T_h$ value is chosen such as the error $E(T_h)$ is minimized. By differentiating $E(T_h)$ with respect to $T_h$ and equating the result to zero, it is possible to use the following equation to define the optimum threshold value $T_h$:

$$AT_h^2 + BT_h + C = 0 \tag{11}$$

considering

$$\begin{aligned} A &= \sigma_h^2 - \sigma_{h+1}^2 \\ B &= 2\cdot(\mu_h\sigma_{h+1}^2 - \mu_{h+1}\sigma_h^2) \\ C &= (\sigma_h\mu_{h+1})^2 - (\sigma_{h+1}\mu_h)^2 + 2\cdot(\sigma_h\sigma_{h+1})^2\cdot\ln\left(\frac{\sigma_{h+1}P_h}{\sigma_h P_{h+1}}\right) \end{aligned} \tag{12}$$

Although the above quadratic equation has two possible solutions, only one of them is feasible, i.e. a positive value falling within the interval.

## 5. Experimental Results

By considering that the mixture parameters are extracted from the fitness function $J$ (Eq.(3)) after applying the ABC algorithm, three experiments are set to evaluate the performance of the proposed algorithm. The first one considers the well-known image "The Camera-man" which is shown by Figure 1a, with its corresponding histogram shown by Figure 1b. The goal is to segment the image in three different pixel classes. According to Eq. (2), during learning, the ABC algorithm adjusts nine parameters, following the minimization procedure conducted by Eq. (3). In this experiment, a population of 20 ($N_p$) bees is considered, with ten employed and ten onlookers bees. Each candidate holds 9 dimensions, such as:

$$I_N = \left\{P_1^N, \sigma_1^N, \mu_1^N, P_2^N, \sigma_2^N, \mu_2^N, P_3^N, \sigma_3^N, \mu_3^N\right\} \tag{13}$$

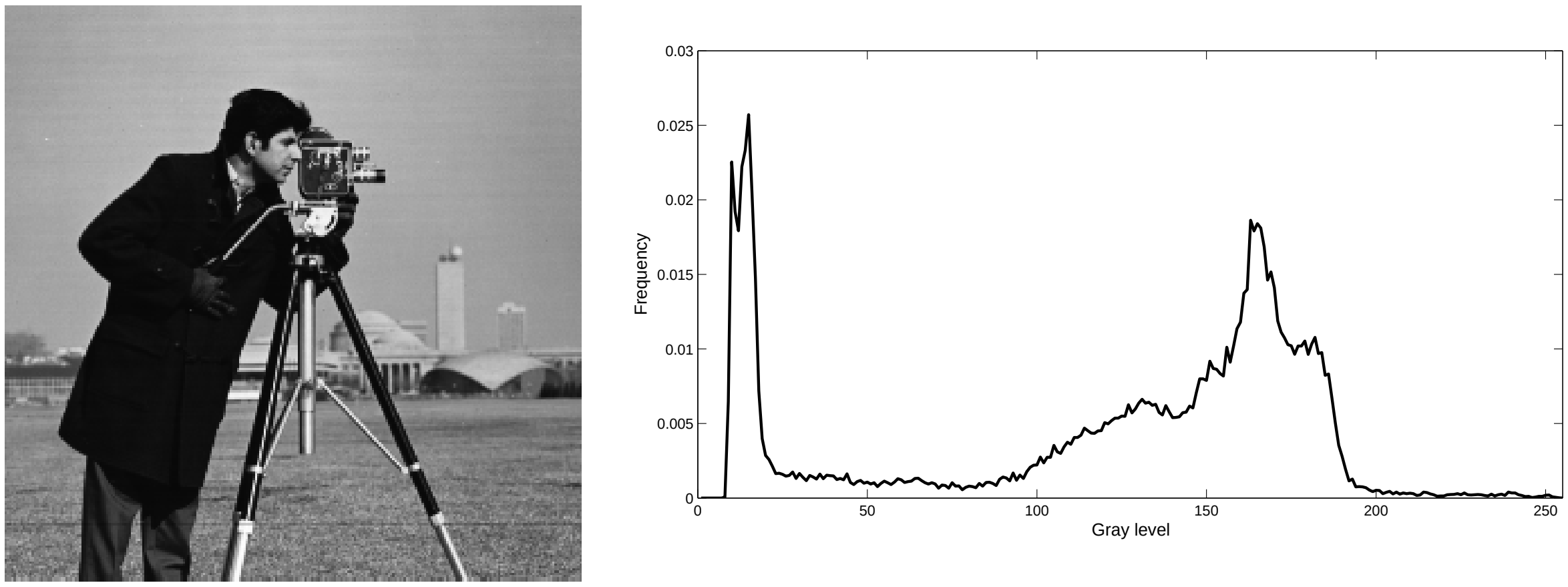

(a) (b)

**Fig. 1.** (a) Original image "The Cameraman", and (b) its correspondent histogram.

with *N* representing the individual's number. The parameters ( $P, \sigma, \mu$ ) are randomly initialized, but assuming some restrictions to each parameter (for example $\mu$ must fall between 0 and 255).

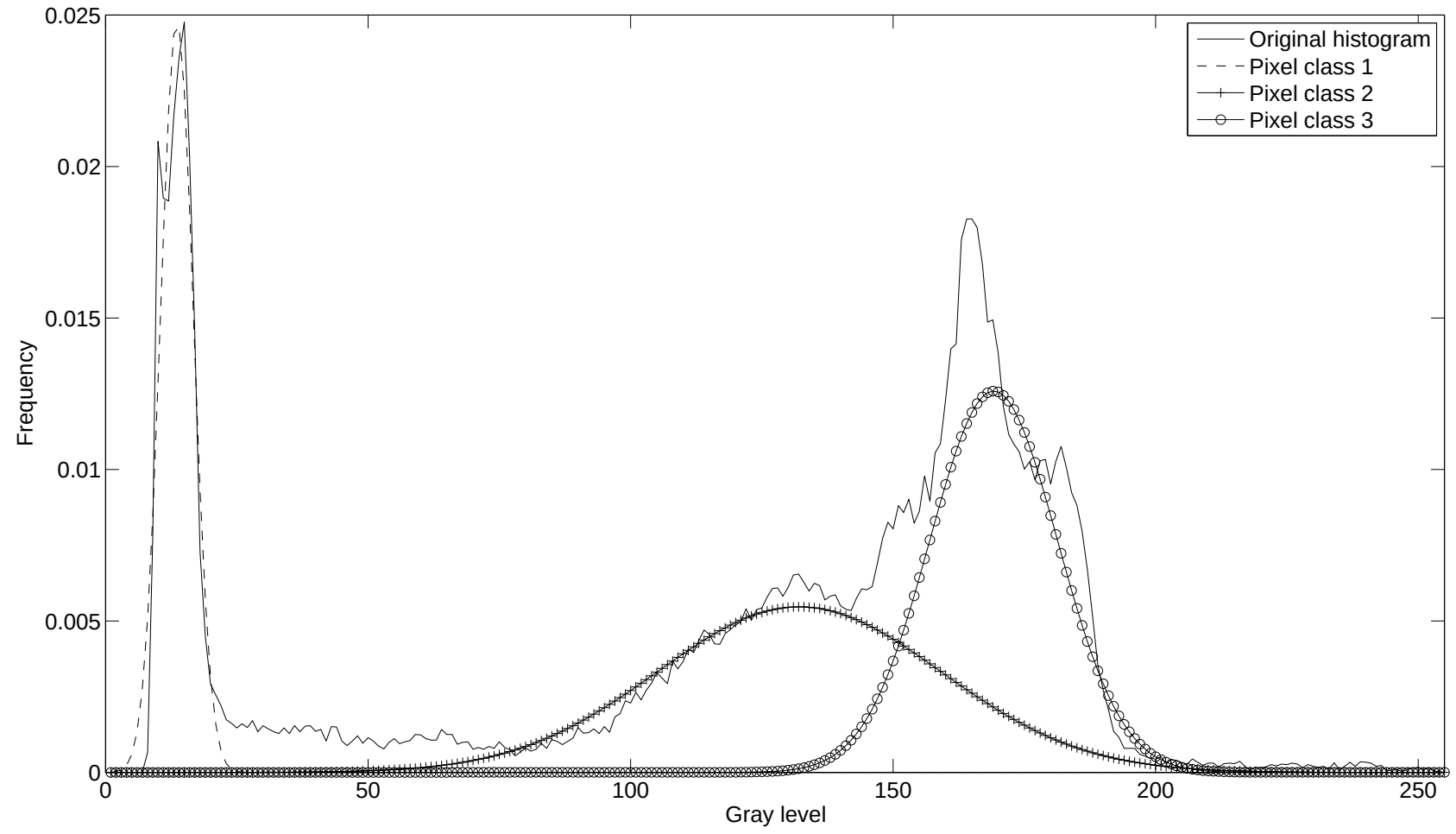

(a)

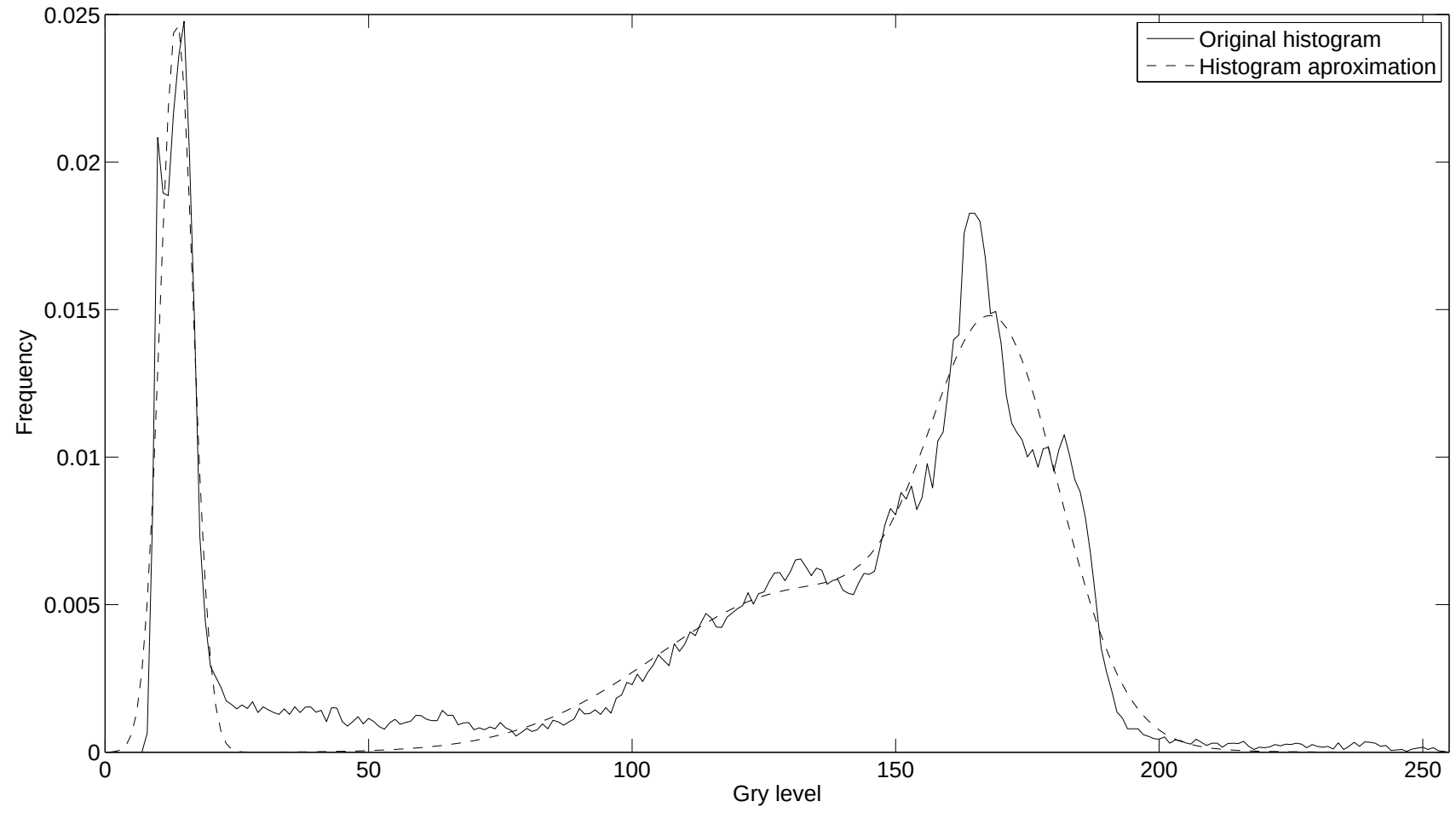

(b)

**Fig. 2.** Applying the ABC algorithm for 3 classes and its results: (a) Gaussian functions for each class, (b) Mixed Gaussian functions (approach to the original histogram).

The experiments suggest that after 200 iterations, the ABC algorithm has converged to the global minimum. Figure 2a shows the obtained Gaussian functions (pixel classes) plotted over the original histogram while Figure 2b shows the Gaussian mixture. Figure 3 shows the segmented image whose threshold values are calculated according to Eqs. (11) and (12).

The algorithm is tested with a greater number of Gaussian functions yielding the need of optimizing more parameters (according to Eq. 3). Thus, twelve parameters are now considered corresponding to the values of four Gaussian functions. One population of 20 bees and 12 dimensions are used for the test. Figure 4a shows the Gaussian functions (pixel classes) plotted over the histogram after 200 iterations, while in Figure 4b presents the Gaussian mixture. The segmented image is depicted by Figure 5.

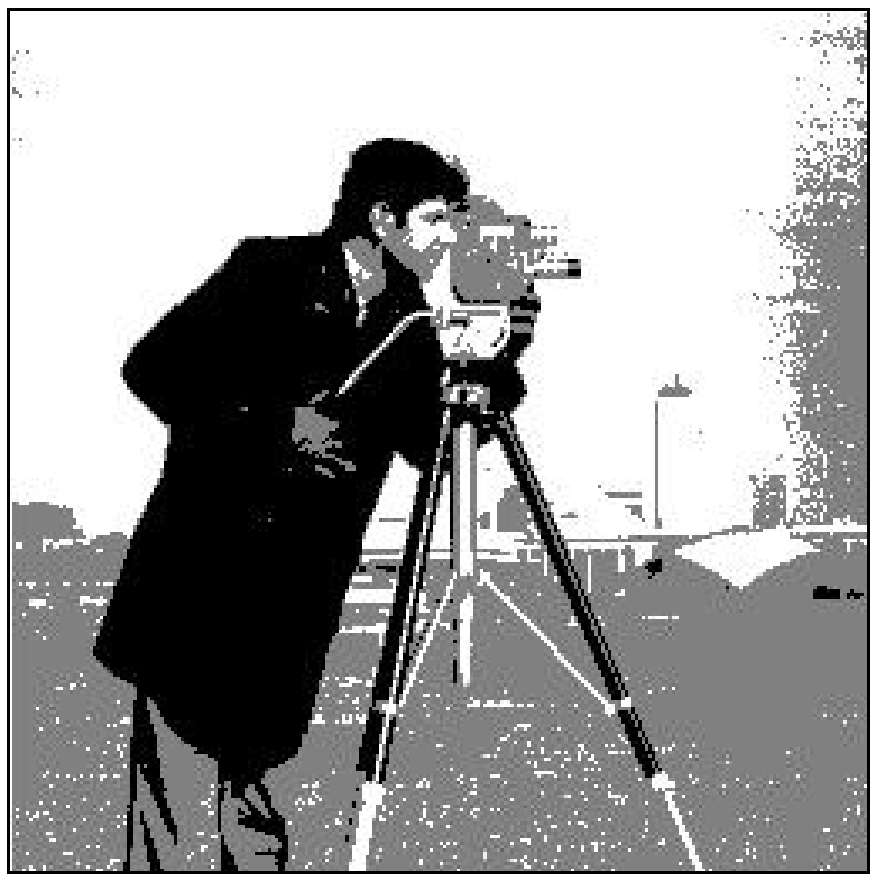

**Fig. 3.** Segmented image considering only three classes.

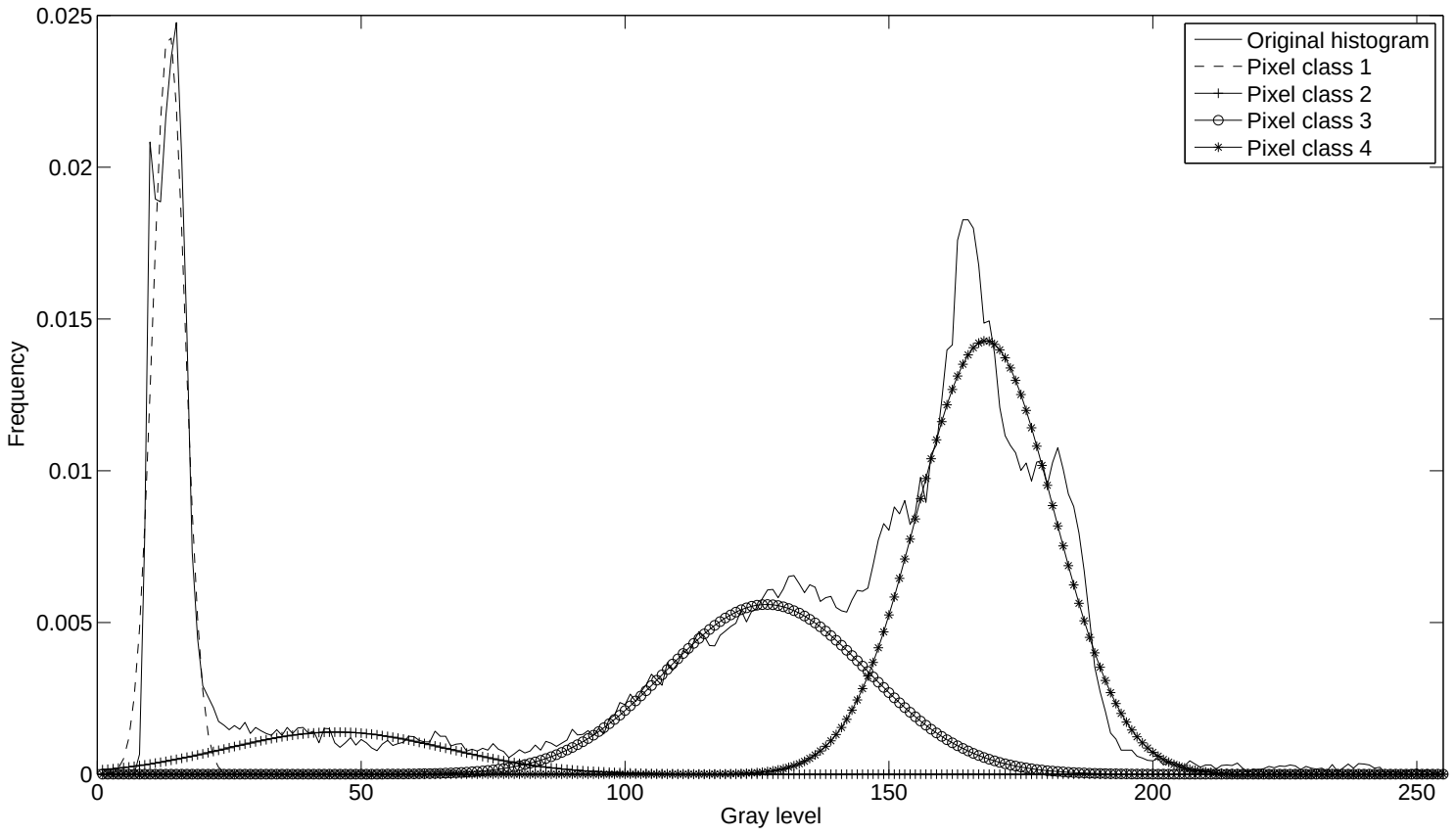

(a)

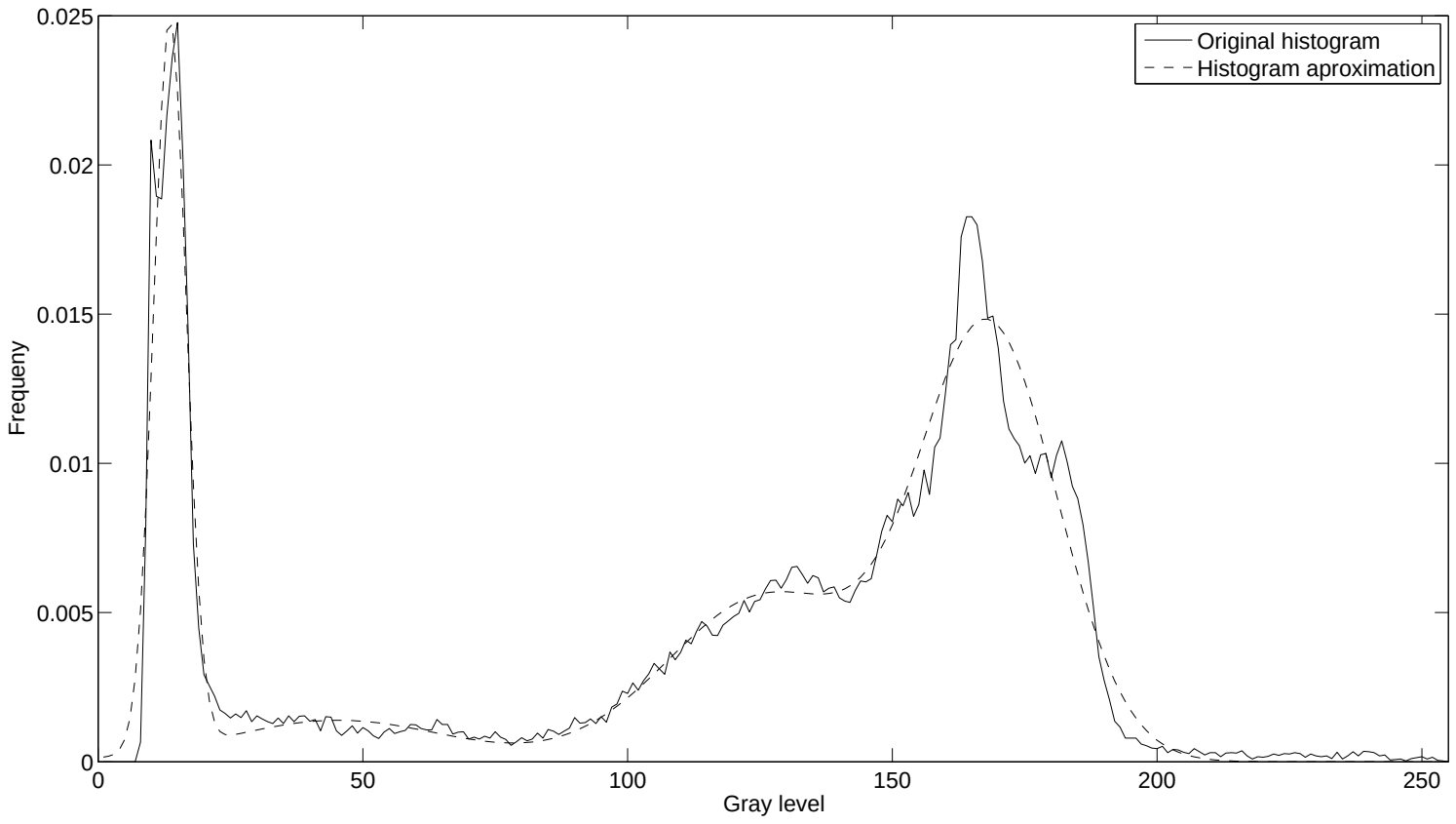

(b)

**Fig. 4.**Segmentation of the test image as it was obtained by the ABC algorithm considering 4 classes: (a) Gaussian functions for each class. (b) Mixed Gaussian functions approaching the original histogram.

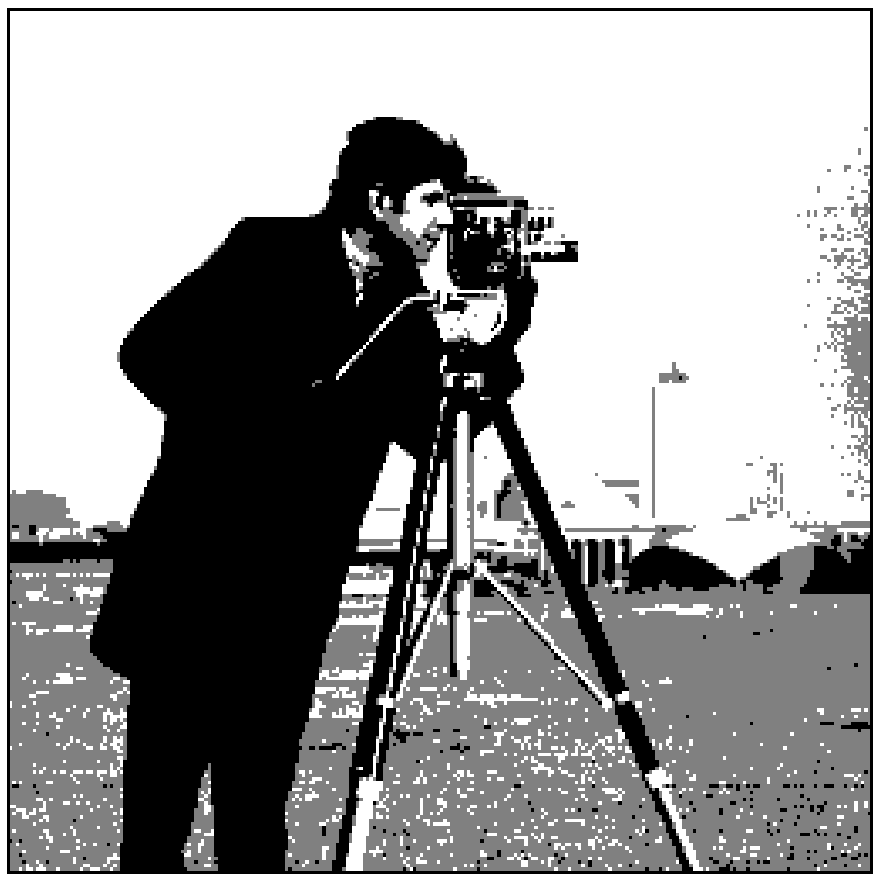

**Fig. 5.** Segmented image considering four classes.

The second experiment considers the popular benchmark image known as "The scene" (see Figure 6a). The image's histogram is presented by Figure 6b. Following the first experiment, the image is segmented considering four pixel classes. The optimization is performed by the ABC algorithm which results in the classes shown by Figure 7a. In turn, Figure 7b presents the Gaussian mixture as it results from the addition of other Gaussian functions. Figure 8 shows the image segmentation considering four classes.

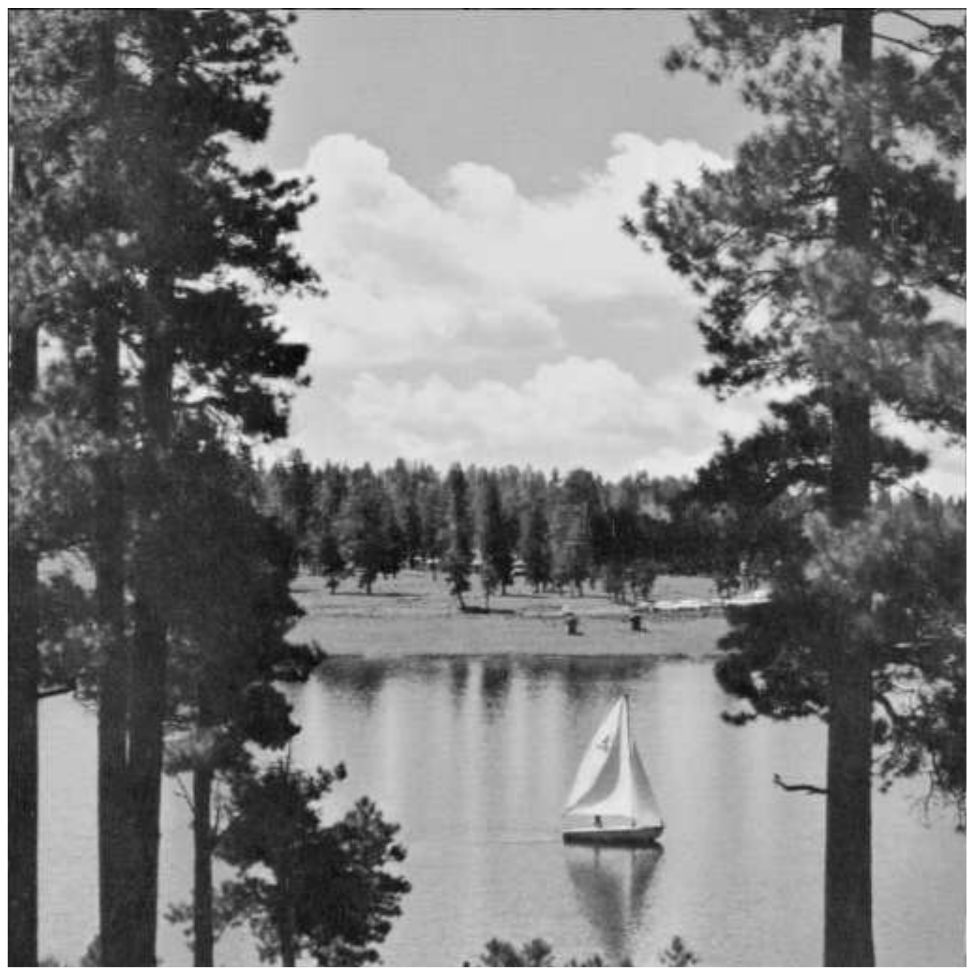

(a)

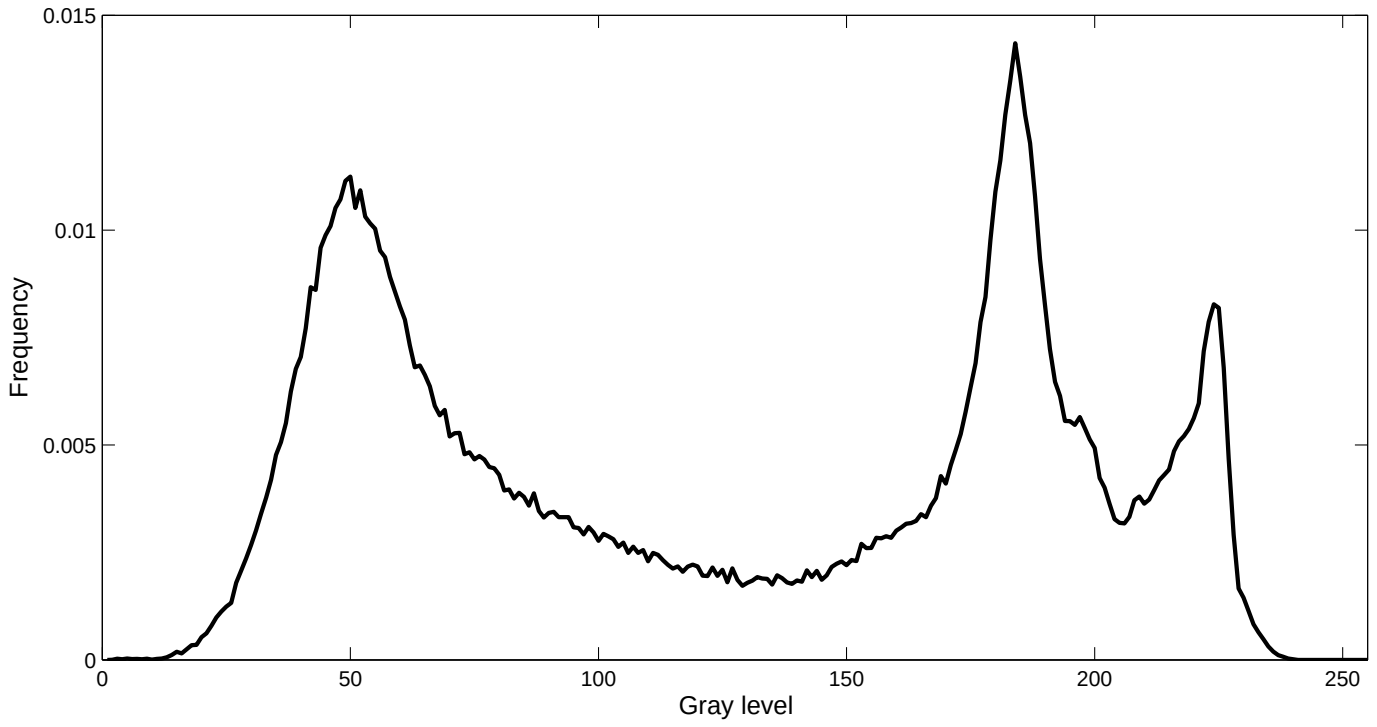

(b)

**Fig. 6.** Second experiment, (a) the original image "The scene", and (b) its histogram.

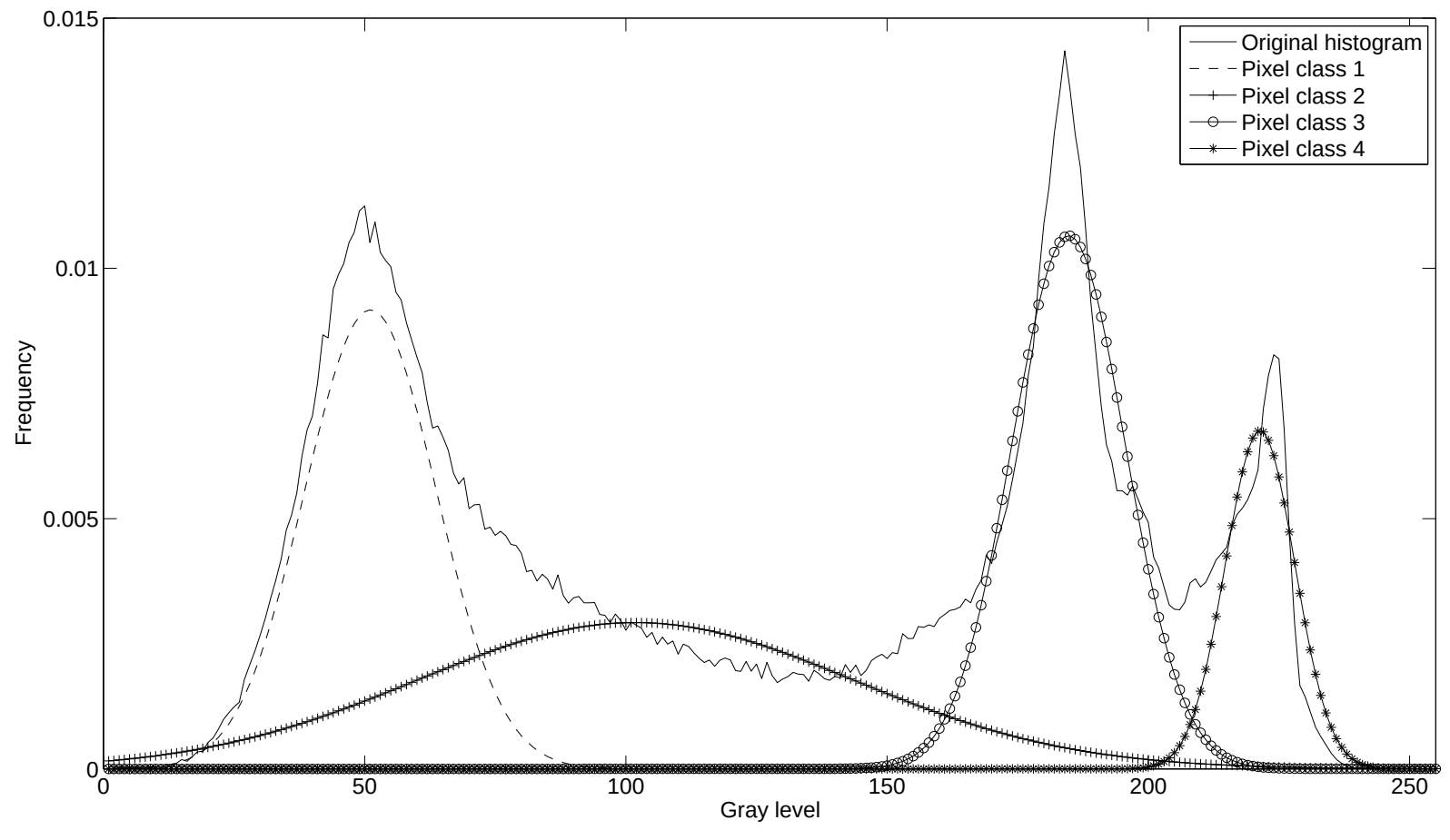

(a)

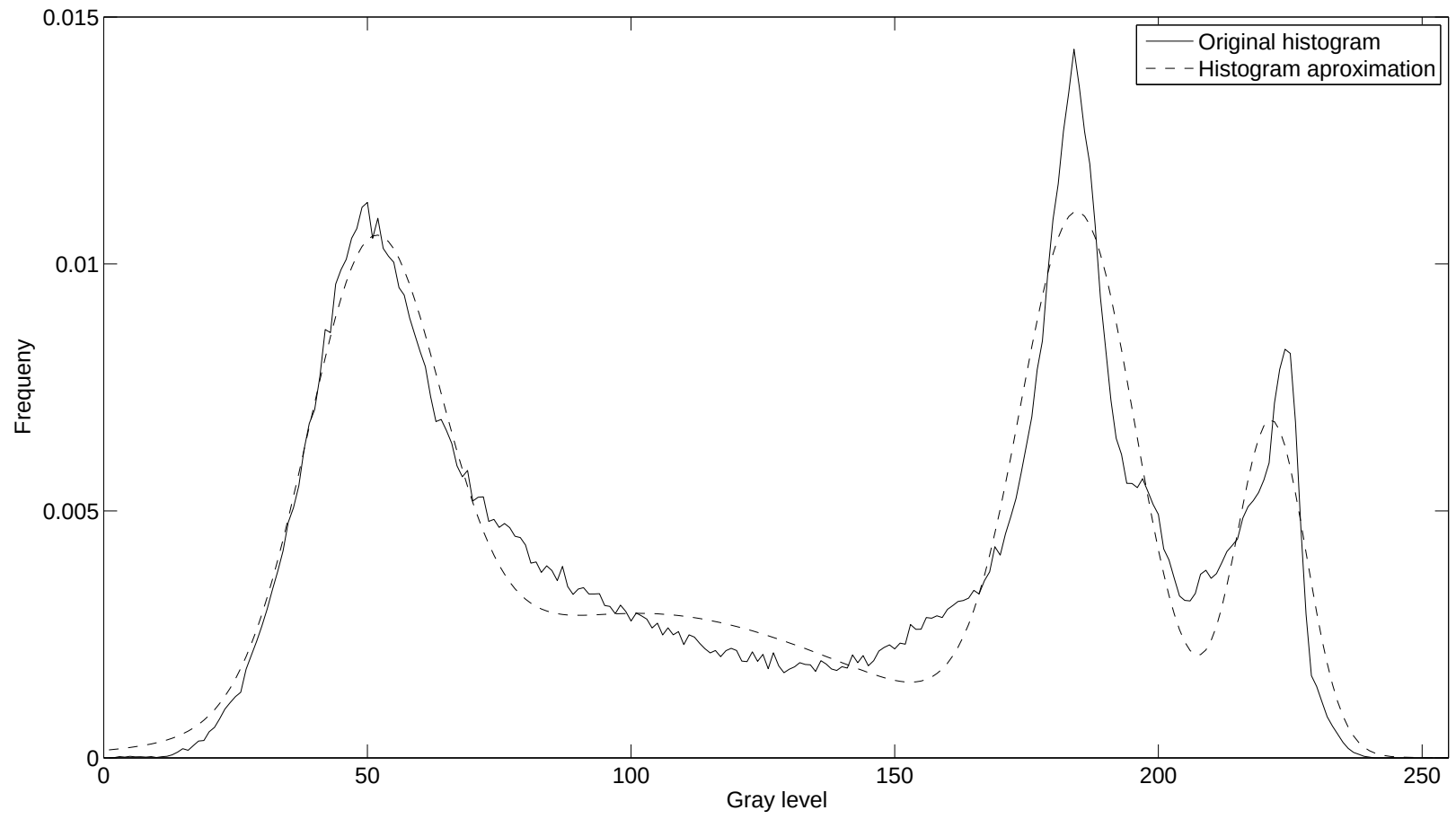

(b)

**Fig. 7.** Results obtained by the ABC algorithm for 4 classes: (a) Gaussian functions at each class, (b) Mixed Gaussian functions approaching the original histogram

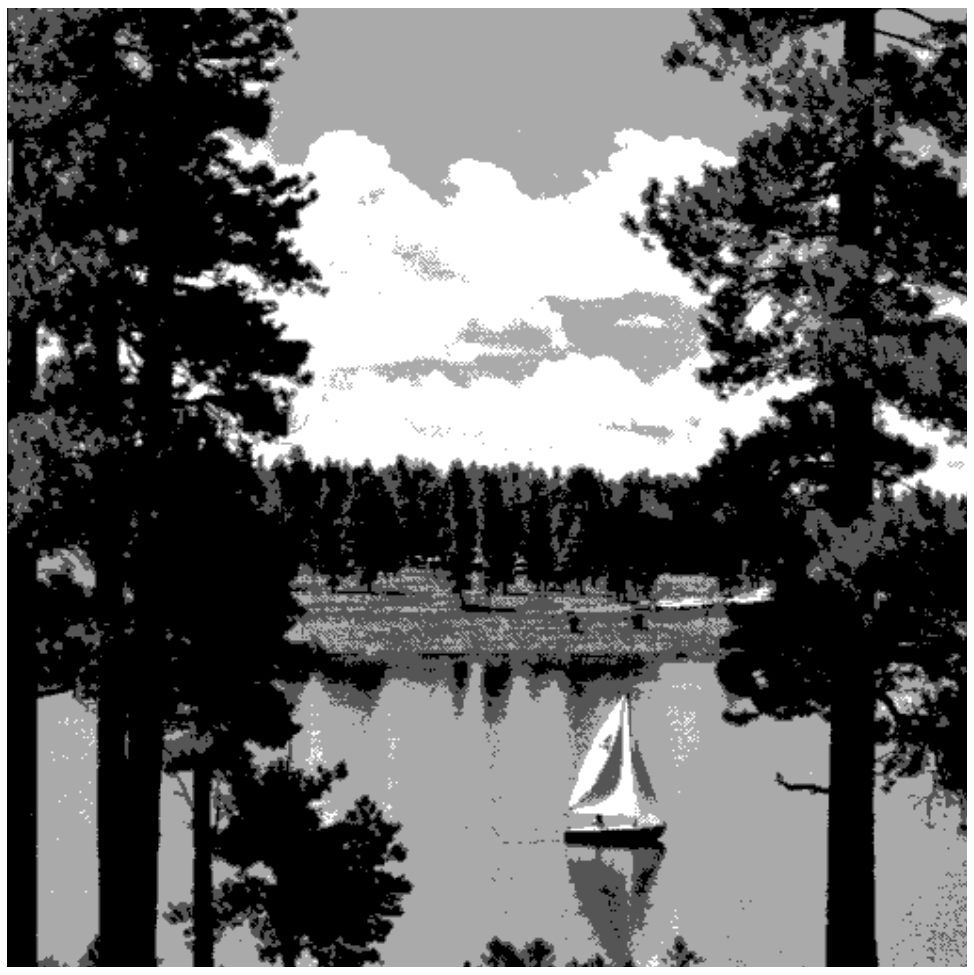

**Fig. 8.** Segmented image considering four classes.

The final experiment considers a blood-smear image as it is processed by the ABC algorithm. The image shows a set of leukocytes cells in the blood smear (darker cells on image of Figure 9a). The gray blobs represent the red blood cells while the background is white yielding only three classes to be considered.

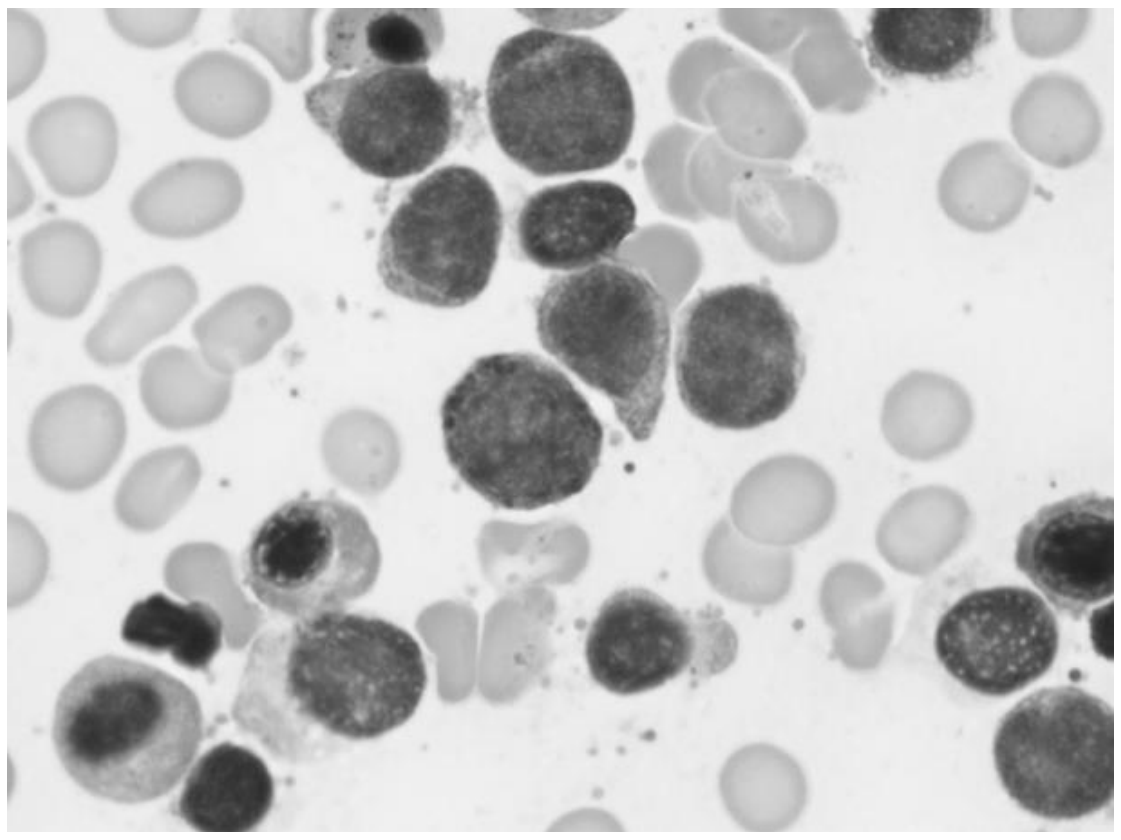

(a)

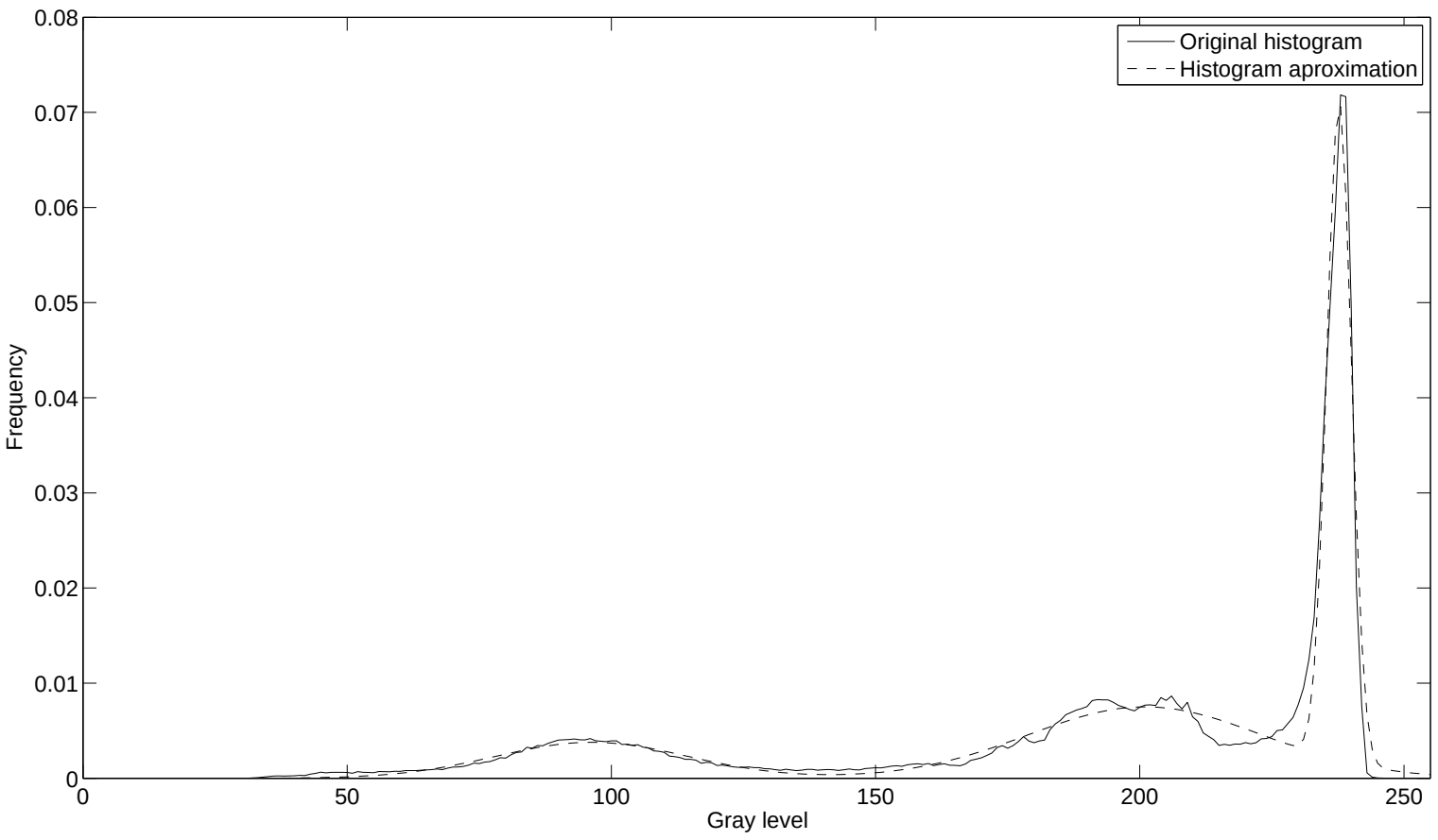

(b)

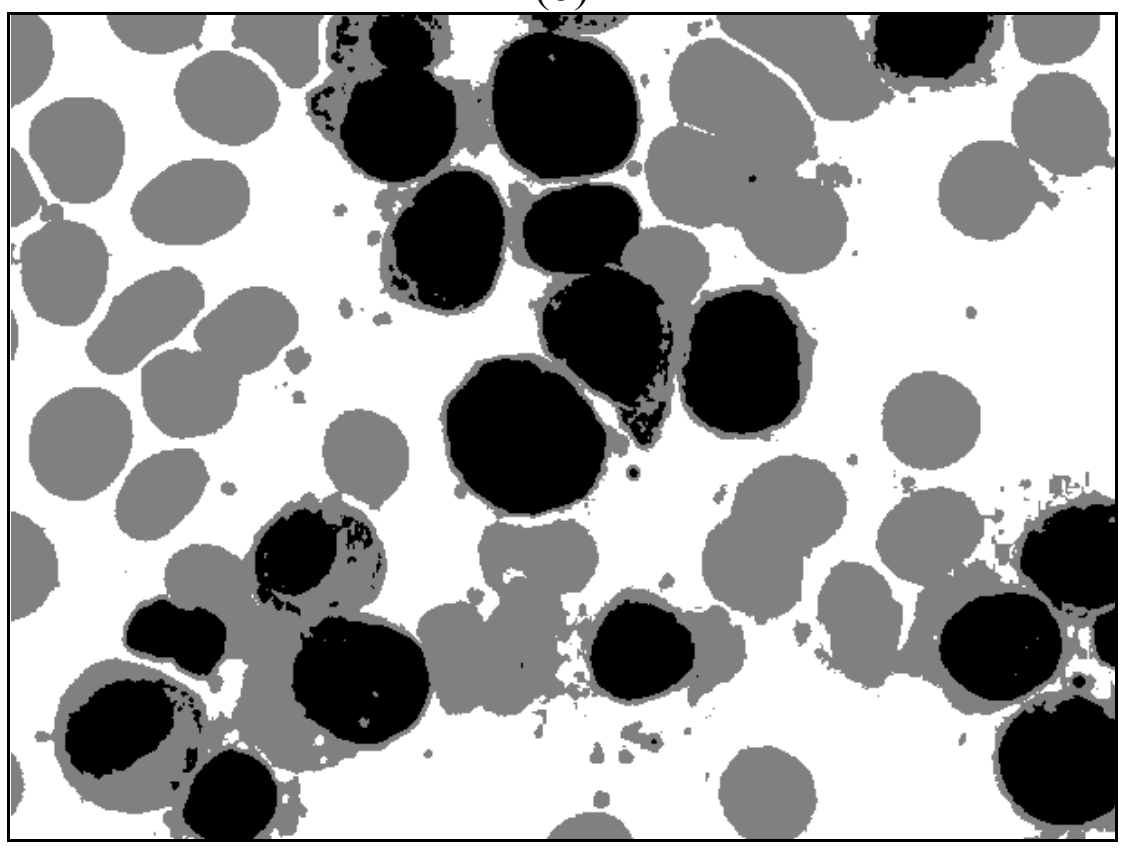

(c)

**Fig. 9.** Segmentation of a blood smear image considering three classes for the ABC algorithm: (a) Original image, (b) Comparison between the original histogram and the Gaussian approach, (c) the segmented image.

*5.1 Comparing the ABC algorithm vs. the EM and LM methods.*

This section discusses on the comparison between ABC and other algorithms such as the EM algorithm and the Levenberg-Marquardt (LM) method which are commonly employed for determining Gaussian mixtures. The discussion focuses on the following issues: sensitivity to the initial conditions, convergence and computational costs.

*a) Sensitivity to initial conditions.* This experiment considers different initial values for all methods assuming the same histogram in the approximation task. After convergence, only final parameters representing the Gaussian mixture are reported. Figure 10a shows the image used in the comparison while Figure 10b pictures the histogram. All experiments are conducted several times in order to assure consistency. Only two different initial states with the highest variation are reported in Table 1. Likewise, Figure 11 shows the obtained segmented images considering two initial conditions as it is reported by Table 1. In the ABC case, the algorithm does not require initialization as random initial values are employed. However, in order to assure a valid comparison, same initial values are considered for the EM, the LM and the ABC method.

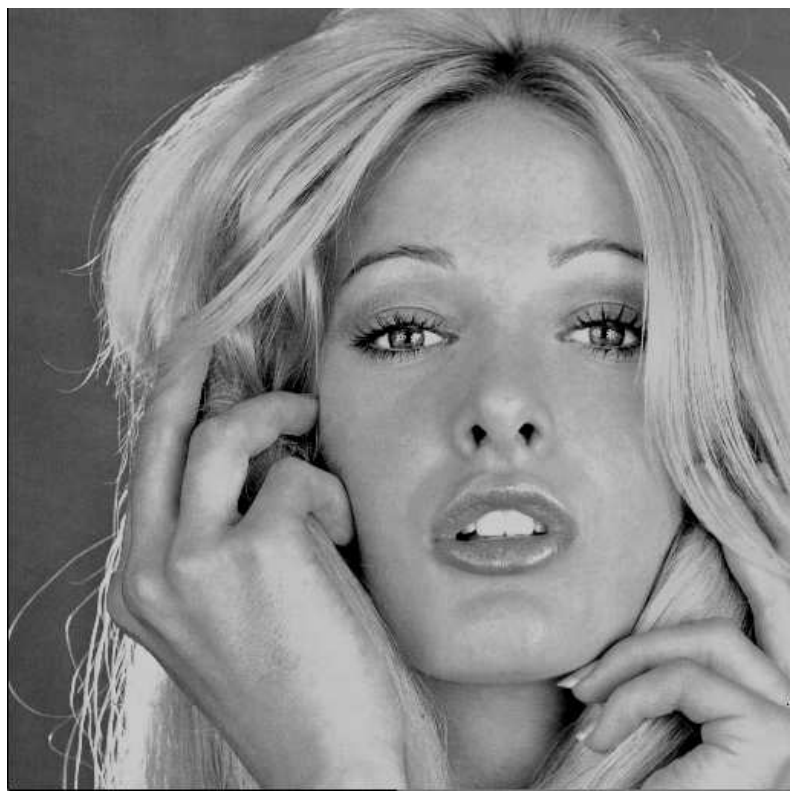

(a)

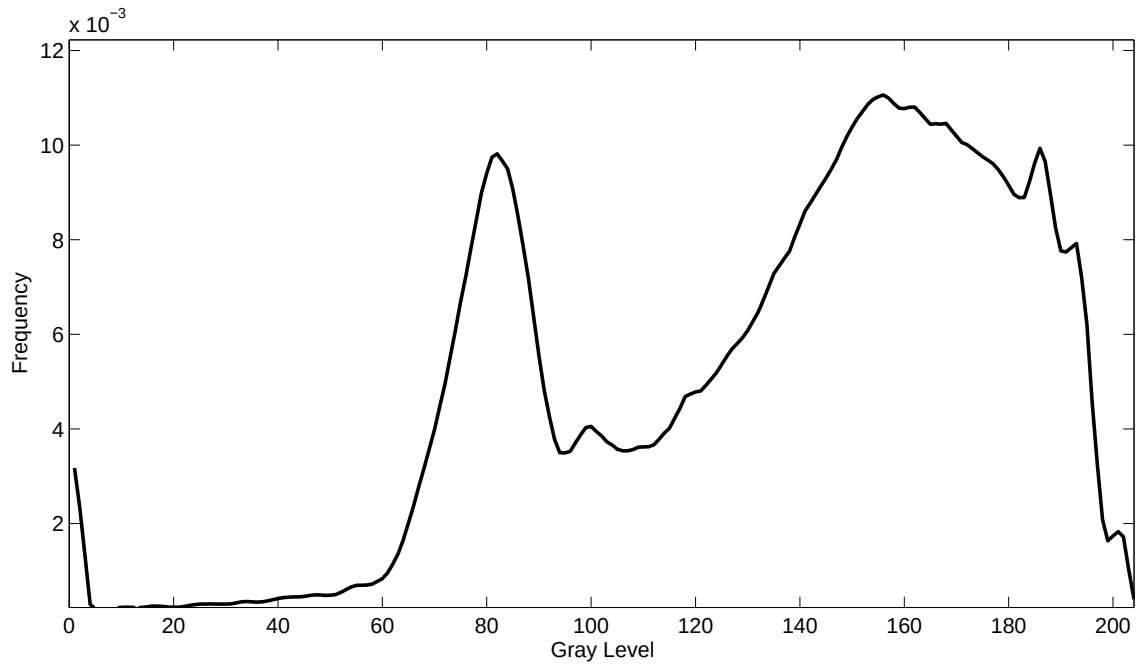

(b)

**Figure 10.** (a) Original image used for the comparison experiment and (b) its corresponding histogram.

| **Parameters** | **Initial condition 1** | **EM** | **LM** | **ABC** | **Initial condition 2** | **EM** | **LM** | **ABC** |
|---|---|---|---|---|---|---|---|---|
| $\mu_1$ | 40.6 | 33.13 | 32.12 | 32.01 | 10 | 20.90 | 31.80 | 32.50 |
| $\mu_2$ | 81.2 | 81.02 | 82.05 | 82.30 | 100 | 82.78 | 80.85 | 82.42 |
| $\mu_3$ | 121.8 | 127.52 | 127 | 127.00 | 138 | 146.67 | 128 | 127.72 |
| $\mu_4$ | 162.4 | 167.58 | 166.80 | 166.10 | 200 | 180.72 | 165.90 | 166.50 |
| $\sigma_1$ | 15 | 25.90 | 25.50 | 25.30 | 10 | 18.52 | 20.10 | 25.01 |
| $\sigma_2$ | 15 | 9.78 | 9.70 | 9.80 | 5 | 12.52 | 9.81 | 10.00 |
| $\sigma_3$ | 15 | 17.72 | 17.05 | 17.71 | 8 | 20.5 | 15.15 | 17.57 |
| $\sigma_4$ | 15 | 17.03 | 17.52 | 17.21 | 22 | 10.09 | 18.00 | 17.22 |
| $P_1$ | 0.25 | 0.0313 | 0.0310 | 0.307 | 0.20 | 0.0225 | 0.0312 | 0.317 |
| $P_2$ | 0.25 | 0.2078 | 0.2081 | 0.201 | 0.30 | 0.2446 | 0.2079 | 0.255 |
| $P_3$ | 0.25 | 0.2508 | 0.2500 | 0.249 | 0.20 | 0.5232 | 0.2502 | 0.260 |
| $P_3$ | 0.25 | 0.5102 | 0.5110 | 0.555 | 0.30 | 0.2098 | 0.5108 | 0.511 |

**Table 1.** Comparison between the EM, the LM and the ABC algorithm, considering two different initial conditions.

By analyzing the information in Table 1, the sensitivity of the EM algorithm to initial conditions becomes evident. Figure 11 shows a clear pixel misclassification in some sections of the image as a consequence of such sensitivity.

**Initial condition set number 1**

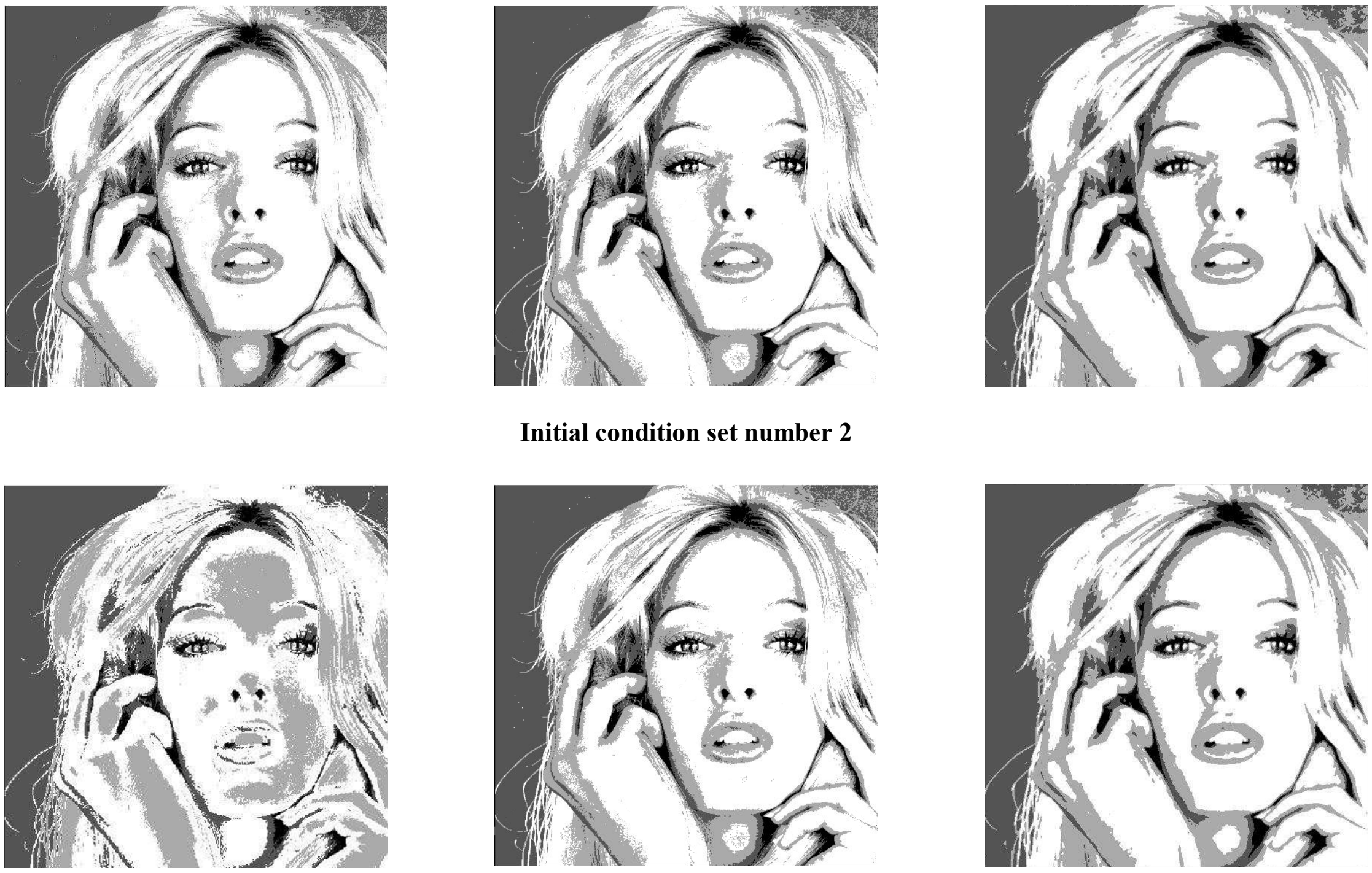

**EM** **LM** **ABC**

**Figure 11.** Segmented images after applying the EM, the LM and the ABC algorithm with different initial conditions.

*b) Convergence and computational cost.* The experiment aims to measure the number of required steps and the computing time spent by the EM, the LM and the ABC algorithm required to calculate the parameters of the Gaussian mixture in benchmark images (see Figure 12a-c). All experiments consider four classes. Table 2 shows the averaged measurements as they are obtained from 20 experiments. It is evident that the EM is the slowest to converge (iterations) and the LM shows the highest computational cost (time elapsed) because it requires complex Hessian approximations. On the other hand, the ABC shows an acceptable trade off between its convergence time and its computational cost. Finally, Figure 12 below shows the segmented images as they are generated by each algorithm.

| **Iterations** / **Time elapsed** | **(a)** | **(b)** | **(c)** |
|---|---|---|---|
| **EM** | 1855 | 1833 | 1870 |
| | 2.72s | 2.70s | 2.73s |
| **LM** | 985 | 988 | 958 |
| | 4.03s | 4.04s | 4.98s |
| **ABC** | 409 | 399 | 512 |
| | 0.78s | 0.70s | 0.81s |

**Table 2.** Iterations and time requirements of the EM, the LM and the ABC algorithm as they are applied to segment benchmark images (see Figure 12).

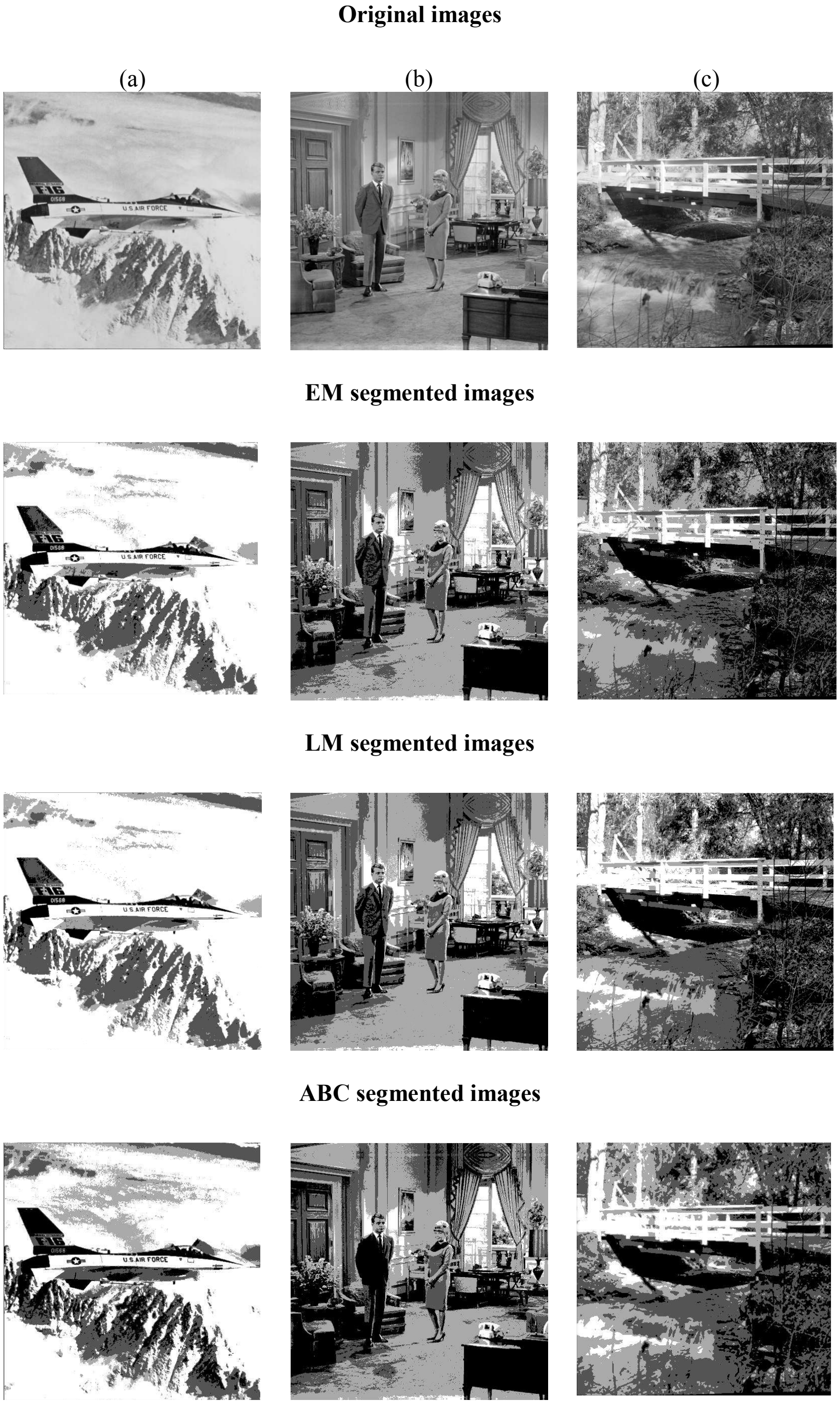

**Figure 12.** Original benchmark images a)-c), and segmented images obtained by the EM, the LM and the ABC algorithms.

## 6. Conclusions

In this paper, an automatic image multi-threshold approach based on the Artificial Bee Colony (ABC) algorithm is proposed. The segmentation process is considered to be similar to an optimization problem. The algorithm approximates the 1-D histogram of a given image using a Gaussian mixture model whose parameters are calculated through the ABC algorithm. Each Gaussian function approximating the histogram represents a pixel class and therefore one threshold point.

Experimental evidence shows that the ABC algorithm has an acceptable compromise between its convergence time and its computational cost when it is compared to the Expectation-Maximization (EM) method and the Levenberg-Marquardt (LM) algorithm. Additionally, the ABC algorithm also exhibits a better performance under certain circumstances (initial conditions) on which it is well-reported in the literature [18] that the EM has underperformed. Finally, the results have shown that the stochastic search accomplished by the ABC method shows a consistent performance with no regard of the initial value and still showing a greater chance to reach the global minimum.